\documentclass{article} % For LaTeX2e
\usepackage{iclr2026_conference,times}

\usepackage{amsmath,amsfonts,bm}

\def\eqref#1{equation~\ref{#1}}
\def\1{\bm{1}}

\DeclareMathAlphabet{\mathsfit}{\encodingdefault}{\sfdefault}{m}{sl}
\SetMathAlphabet{\mathsfit}{bold}{\encodingdefault}{\sfdefault}{bx}{n}

\usepackage{hyperref}
\usepackage{url}
\usepackage[utf8]{inputenc}
\usepackage{algorithm}
\usepackage{algorithmic}

\usepackage{graphicx}
\usepackage{caption}
\usepackage{booktabs}
\usepackage{siunitx}     % 用于数值列对齐
\usepackage{multirow}    % 如果表格有跨行单元格
\usepackage{array}       % 增强表格列格式控制
\usepackage{pifont}

\usepackage{bm} % 粗体数学符号（可选）
\usepackage{amsthm}

\usepackage{subcaption}

\title{Dynamic-TreeRPO: Breaking the Independent Trajectory Bottleneck with Structured Sampling}

\author{
\textbf{Xiaolong Fu}\textsuperscript{1}\thanks{Equal contribution.} \quad
\textbf{Lichen Ma}\textsuperscript{1}\footnotemark[1] \quad
\textbf{Zipeng Guo}\textsuperscript{1,2}\footnotemark[1] \quad
\textbf{ShiPing Dong}\textsuperscript{1,4} \quad
\textbf{Lan Yang}\textsuperscript{5} \\
\textbf{Tan Lit Sin}\textsuperscript{1,3} \quad
\textbf{Gaojing Zhou}\textsuperscript{1} \quad
\textbf{Yu He}\textsuperscript{1} \quad
\textbf{Jingling Fu}\textsuperscript{1} \quad
\textbf{Shizhe Zhou}\textsuperscript{4} \\
\textbf{Junshi Huang}\textsuperscript{1}\thanks{Corresponding Author.} \quad
\textbf{Yan Li}\textsuperscript{1} \quad \\
\textsuperscript{1}JD.COM \quad
\textsuperscript{2}Sun Yat-sen University \quad
\textsuperscript{3}Tsinghua University \quad \\
\textsuperscript{4}Hunan Univerisity \quad
\textsuperscript{5}Beijing University of Chemical Technology \quad
}

\iclrfinalcopy % Uncomment for camera-ready version, but NOT for submission.
\makeatletter
\def\@oddhead{}
\def\@evenhead{}
\makeatother

\begin{document}
\maketitle

\begin{abstract}

The integration of Reinforcement Learning (RL) into flow matching models for text-to-image (T2I) generation has driven substantial advances in generation quality.
However, these gains often come at the cost of exhaustive exploration and inefficient sampling strategies due to slight variation in the sampling group.
Building on this insight, we propose Dynamic-TreeRPO, which implements the sliding-window sampling strategy as a tree-structured search with dynamic noise intensities along depth.
We perform GRPO-guided optimization and constrained Stochastic Differential Equation (SDE) sampling within this tree structure. By sharing prefix paths of the tree, our design effectively amortizes the computational overhead of trajectory search.
With well-designed noise intensities for each tree layer, Dynamic-TreeRPO can enhance the variation of exploration without any extra computational cost.
Furthermore, we seamlessly integrate Supervised Fine-Tuning (SFT) and RL paradigm within Dynamic-TreeRPO to construct our proposed LayerTuning-RL, reformulating the loss function of SFT as a dynamically weighted Progress Reward Model (PRM) rather than a separate pretraining method.
By associating this weighted PRM with dynamic-adaptive clipping bounds, the disruption of exploration process in Dynamic-TreeRPO is avoided.
Benefiting from the tree-structured sampling and the LayerTuning-RL paradigm, our model dynamically explores a diverse search space along effective directions.
Compared to existing baselines, our approach demonstrates significant superiority in terms of semantic consistency, visual fidelity, and human preference alignment on established benchmarks, including HPS-v2.1, PickScore, and ImageReward. In particular, our model outperforms SoTA by $4.9\%$, $5.91\%$, and $8.66\%$ on those benchmarks, respectively, while improving the training efficiency by nearly $50\%$.

\end{abstract}

\section{Introduction}
Flow matching-based image generation models ~\citep{lipman2022flow, liu2022flow, esser2024scaling, flux2024, batifol2025flux}, renowned for their solid theoretical foundations and impressive performance, have demonstrated remarkable results in text-to-image tasks. However, significant challenges remain in scenarios involving text rendering, numerals, and fine-grained attribute control. Recent advances ~\citep{ouyang2022training, fan2023dpok, xu2023imagereward, wallace2024diffusion, gong2025onereward} have shown that incorporating GRPO~\citep{shao2024deepseekmath} in the pretraining phase can lead to improved performance. 
However, the inherent trial-and-error nature of reinforcement learning fundamentally limits the efficiency and effectiveness of these approaches.

\begin{figure*}[t]
    \centering
    \includegraphics[width=0.99\textwidth]{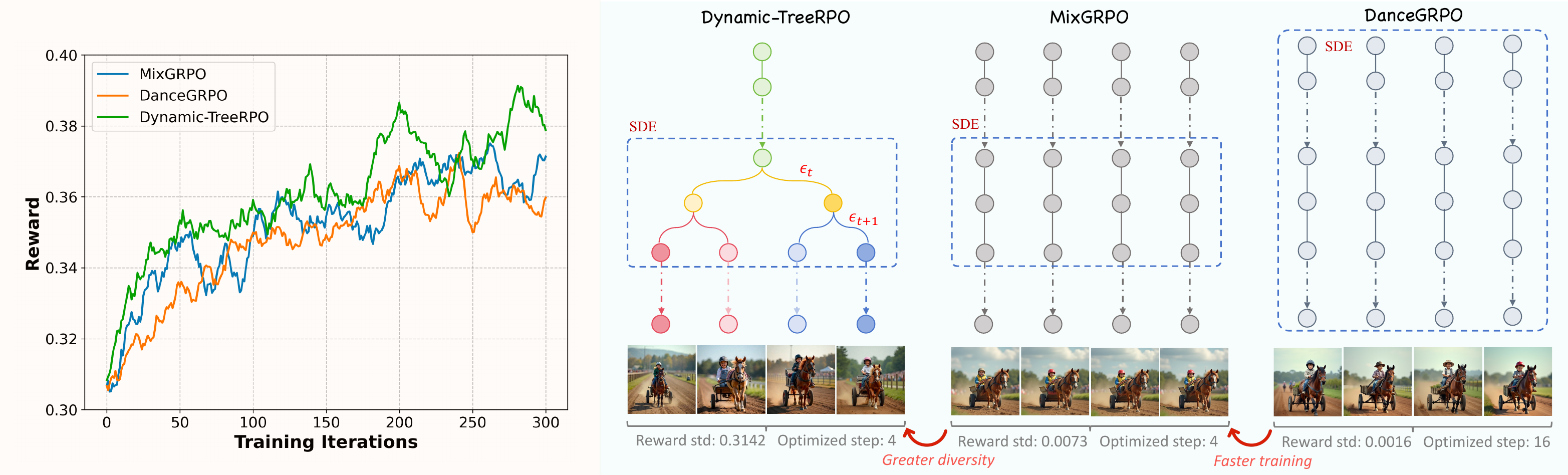}
    \caption{Compare with the previous method. $Left$: The reward curve during training shows that Dynamic-TreeRPO converges more rapidly than both DanceGRPO and MixGRPO, and ultimately achieves significantly better results than either of them. $Right$: Visualization of the different structures. Dynamic-TreeRPO employs a tree structure with a sliding window mechanism. MixGRPO utilizes a sliding window structure, where SDE is applied only during the sliding window period. In contrast, DanceGRPO applies SDE throughout the entire process.}
    \label{fig:compare}
\end{figure*}

Current GRPO-based probability flow models~\citep{liu2025flow, xue2025dancegrpo, li2025mixgrpo, he2025tempflow} introduce stochasticity at each time step via stochastic differential equations, and leverage GRPO to optimize the entire state-action sequence. 
However, these approaches incur substantial computational overhead during the exploratory denoising process, significantly slowing down the training speed. 
Subsequent methods such as MixGRPO~\citep{li2025mixgrpo} and Tempflow-GRPO~\citep{he2025tempflow} accelerate training by reducing the number of SDE sampling in denoising process, leading to slight variation and similar trajectories in sampling group.
Meanwhile, the prevailing approaches ~\citep{bai2025qwen2, chu2025sft, zhang2025reasongen} still adhere to the sequential SFT-then-RL paradigm. 
This fully decoupled two-stage setup is susceptible to catastrophic forgetting, inefficient exploration, and hallucinations ~\citep{lv2025towards}.
To address these challenges, we propose Dynamic-TreeRPO, an intuitive and effective solution that integrates a hybrid ODE-SDE strategy into the framework composed of a sliding window and a tree structure.
Our approach enables diversified sample generation with shared prefixes of sampling trajectories.
Specifically, we implement the SDE sampling strategy within the sliding window as tree structure and employ ODE sampling elsewhere, thereby confining the stochasticity in this tree.
The common inference steps of the ancestor nodes in the tree are computed once and shared for all descendants, pursing for efficient computation.
To expand the exploration of tree-structured sampling, we design a dynamic noise intensity along the depth of the tree.
As observed in Figure ~\ref{fig:compare}, the reward variance of Dynamic-TreeRPO is larger than that of other methods, indicating a broader exploration space of our method.
% However, shared ancestors may harm the diversity of trajectories, leading to limited exploration space.
% Inspired by the observation that 

Furthermore, we introduce a novel training strategy, denoted LayerTuning-RL, to seamlessly integrate SFT with Dynamic-TreeRPO.
SFT primarily learns by exploiting high-quality expert data, while RL explores through interaction and feedback from the environment. The conventional approach combines these two by first performing SFT and then RL. 
Although intuitive, this often yields suboptimal results in practice, as the RL phase consistently suffers from issues such as policy hacking and catastrophic forgetting ~\citep{chen2025beyond, zhang2025policy}. 
% The underlying reason is the disconnection created by RL in terms of objective functions, representation spaces, and exploration mechanisms. 
The underlying reason is the disconnection between SFT and RL in the objective function, representation space, and exploration mechanism. 
Generally, the RL stage lacks the capability to continuously anchor to the knowledge acquired in SFT.

To address this, we rethink SFT not as an independent training method, but as a dynamically weighted auxiliary objective within the RL process. 
We apply the modified SFT function at each layer of the tree structure, armed with which the Dynamic-TreeRPO can perform more efficiently and robustly for the exploration of diverse sampling. 
In this way, the SFT function acts as a PRM to guide the sampling direction of each decision node.
% With the introduction of LayerTuning-RL, GRPO can efficiently and stably explore more diverse outputs. 
Additionally, we decouple the clipping bounds in the GRPO algorithm and dynamically adjust them based on the training steps, which prevents entropy collapse and fully leverages the potential of GRPO.
% Our approach constructs a reinforcement learning framework in which expert behavior constraints consistently guide every decision node.
In summary, our contributions are as follows.
\begin{itemize}
    \item We propose Dynamic-TreeRPO, a novel RL training framework that formulates the sampling process as a tree-structured search. 
    % By employing the sliding window strategy and a dynamic noise mechanism, our method amortizes the computational cost across shared tree prefixes, while restricting SDE sampling and GRPO-guided optimization to the tree structure.
    By employing the sliding window strategy and a tree-structured search with dynamic noise mechanism, our method amortizes the computational cost across shared tree prefixes for the purpose of efficient training.
\end{itemize}
\begin{itemize}
    \item We introduce a new training paradigm, LayerTuning-RL, which seamlessly integrates SFT and the proposed Dynamic-TreeRPO. Rather than treating SFT as an isolated training phase, we reformulate it as a dynamically weighted auxiliary objective throughout the training process, effectively mitigating issues such as catastrophic forgetting, inefficient exploration, and model hallucinations.
\end{itemize}
\begin{itemize}
    \item Our method improves the training efficiency of T2I task by nearly $50\%$ over the prior state-of-the-art and achieves superior performance on several benchmarks in terms of semantic consistency, visual quality, and human preference alignment.
\end{itemize}

\section{Related Work}
Diffusion models ~\citep{ho2020denoising,song2020score,song2020denoising,lu2025dpm,lu2022dpm,zheng2023dpm,zhao2023unipc,salimans2022progressive} gradually add noise to data until it becomes random noise, then learn to reverse this process. Sampling is performed using either discrete DDPM steps or probabilistic flow SDE solvers to generate high-fidelity outputs. Flow matching~\citep{lipman2022flow,yin2024one,gao2024diffusion} constructs a continuous path between the noise and data distributions by directly matching the velocity fields, enabling the learning of a continuous time-normalized flow, so that only a few ODE steps are sufficient for deterministic sampling. Recent works, e.g., Flow-GRPO ~\citep{liu2025flow} and Dance-GRPO ~\citep{xue2025dancegrpo}, introduce GRPO into flow matching models, converting deterministic flow models into equivalent SDEs via ODE-to-SDE conversion strategies, while preserving the marginal distributions of the original models to support RL-based stochastic sampling. However, previous methods, e.g., Flow-GRPO~\citep{liu2025flow}, still suffer from inefficiency issues, as they require sampling and optimization over all denoising steps. MixGRPO~\citep{li2025mixgrpo} addresses this problem by proposing a mixed sampling strategy and introducing a sliding window mechanism, where SDE sampling and GRPO-guided optimization are only applied within the window. TempFlow-GRPO~\citep{xue2025dancegrpo} introduces a trajectory branching mechanism that provides process rewards at designated branching points to enable precise credit assignment without requiring dedicated intermediate reward models. Nevertheless, these methods still face challenges such as insufficient intra-group diversity and high computational overhead for group operations. To address these issues, we propose an elegant and intuitive solution, Dynamic-TreeRPO, which reconstructs the processes of different samples into a binary tree with shared prefixes and a sliding window mechanism, based on a hybrid ODE-SDE strategy.

Previous methods~\citep{chu2025sft,chen2025sft} have explored the differences between SFT and RL. ~\citep{chu2025sft} observed that in both visual and rule-based textual environments, RL trained with outcome-based rewards achieves better generalization, while SFT tends to specialize to the specific data distribution, exhibiting memorization of the training data. Consequently, some recent works have focused on hybrid training paradigms to harness their complementary benefits. In the field of LLMs, a two-stage approach~\citep{chen2025beyond} has been introduced into the training process, which leverages the synergy between RL and SFT to enhance model performance. For T2I task, SimpleAR~\citep{wang2025simplear} applies SFT to enhance fidelity and instruction-following capability, and then uses RL to further refine multimodal alignment and mitigate bias. However, this completely decoupled two-stage approach (SFT followed by RL) tends to forget the knowledge acquired during the SFT phase and can lead to inefficiencies in the exploration space of RL. To address this, we are the first to propose a novel training strategy, LayerTuning-RL, for flow-matching-based image generation tasks, which integrates supervised fine-tuning with Dynamic-TreeRPO.

\begin{figure*}[h]
    \centering
    \includegraphics[width=0.99\textwidth]{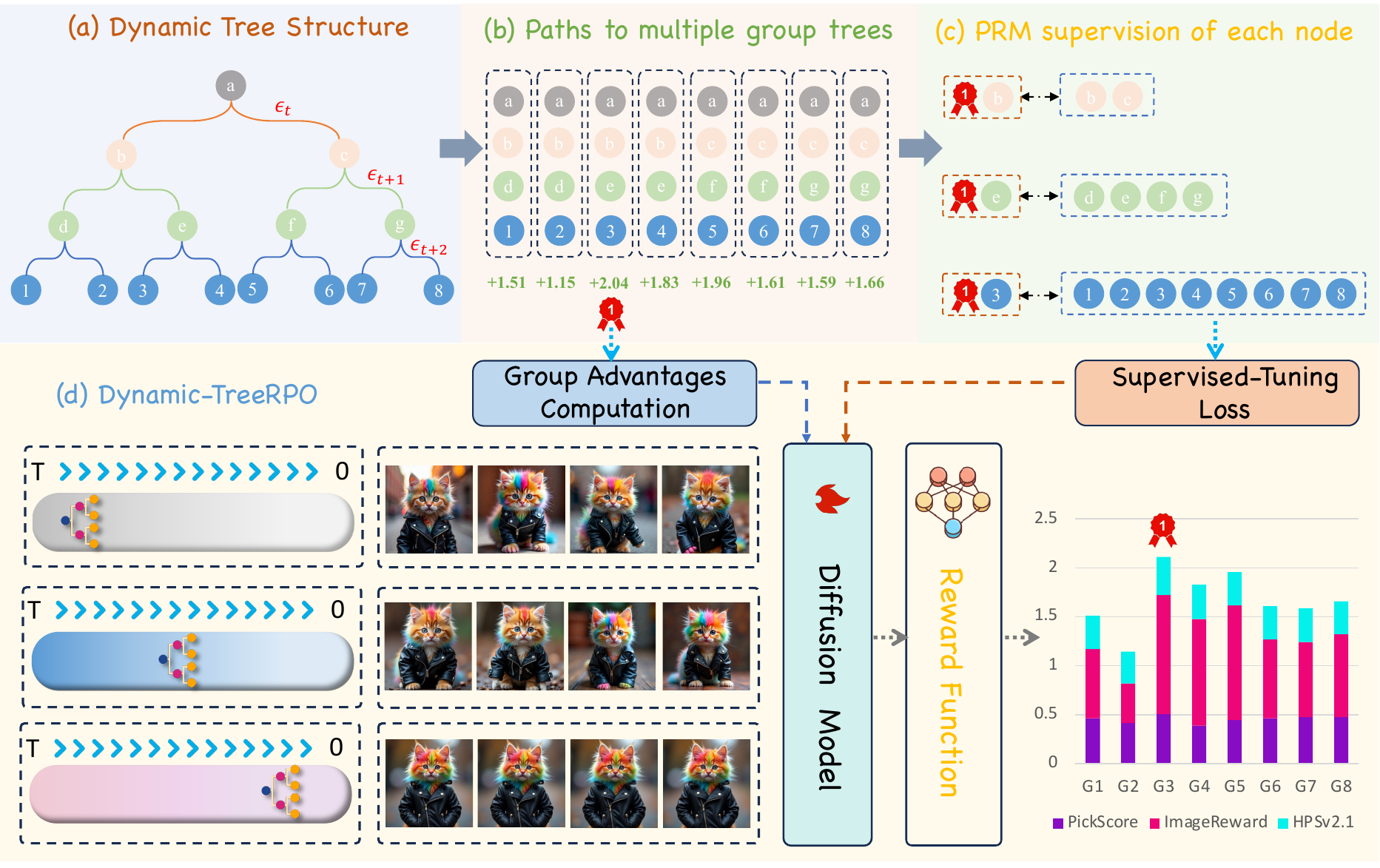}
    \caption{The framework of Dynamic-TreeRPO. $(a)$ Dynamic Tree Structure. Noise intensity is dynamically introduced for the nodes of each layer in the tree structure. $(b)$ Paths to multiple group trees. For each path, the highest reward score is selected. $(c)$ PRM supervision of each node. The node with the maximum reward is used to supervise the model’s predictions at each layer. $(d)$ Training procedure of Dynamic-TreeRPO.}
    \label{fig:main}
\end{figure*}

\section{Method}
In this paper, we aim to establish an efficient, stable, and robust RL paradigm for T2I generation. 
By equipping the sliding window strategy with tree structure sampling, we enable prefix sharing among tree branches to amortize computational overhead.
% , and apply SDE sampling only within the tree.
% Furthermore, we propose a novel training paradigm, LayerTuning-RL, which integrates SFT to alleviate the inefficiency of existing approaches while enhancing the diversity of exploration by amplifying the differences among different tree branches.
Meanwhile, we propose a novel training paradigm, \textit{i.e.} LayerTuning-RL, to integrate SFT function as PRM for the stable training of RL paradigm.
In Section 3.1, we first introduce the main idea of GRPO on flow matching model. Then, we study the weaknesses of GRPO, such as inefficiency and vulnerability, to reward hacking, and propose the Dynamic-TreeRPO in Section 3.2.
In Section 3.3, we describe the dynamic optimization of the upper and lower clipping ranges for GRPO.
Finally, we present the framework of our LayerTuning-RL in Section 3.4.

\subsection{Preliminary: Flow-based GRPO}
In this section, we introduce the core concept of GRPO, and then review how flow-based GRPO converts a deterministic ODE sampler into an SDE sampler with the same marginal distribution to meet the stochastic exploration requirements of GRPO.

\textbf{GRPO on Flow Matching}. RL aims to learn a policy that maximizes the expected cumulative reward by optimizing the policy model to maximize the following objective:
\begin{equation}
% \begin{split}
\mathcal{J}(\theta) = \mathbb{E}_{c \sim C,\substack{\{\mathbf{x}_i\}_{i=1}^G \sim \pi_{\theta_{\text{old}}}(\cdot|c)}} \bigg[\frac{1}{G} \sum_{i=1}^G \frac{1}{T} \sum_{t=1}^T \min\bigg( \mathbf{\rho}_{t,i} A_i, \text{clip}\big( \mathbf{\rho}_{t,i}, 1-\varepsilon, 1+\varepsilon \big) A_i \bigg) \bigg],
% \end{split}
\end{equation}
where $\mathbf{\rho}_{t,i} = \frac{\pi_{\theta}(\mathbf{x}_{t-1,i}|\mathbf{x}_{t,i},c)}{\pi_{\theta_{old}}(\mathbf{x}_{t-1,i}|\mathbf{x}_{t,i},c)}, $ $T$ is the timestep, and $\pi_{\theta}(\mathbf{x}_{t,i-1}|\mathbf{x}_{t,i},c)$ is the policy function used to generate the output at time step $t$ in a Markov Decision Process (MDP). $\varepsilon$ is a hyperparameter. $A_i$ denotes the advantage function. Given a prompt $\mathbf{c}$, the model samples a set of $G$ images $\{\mathbf{x}_{0,i}\}_{i=1}^G$, and obtains the corresponding rewards $R = \{r_1, r_2, \ldots, r_G\}$, which are then standardized as: 
\begin{equation}
A_i=\frac{r_i-\mathrm{mean}(\{r_1,r_2,\cdots,r_G\})}{\mathrm{std}(\{r_1,r_2,\cdots,r_G\})}.
\end{equation}
\textbf{Convert ODE to SDE.} GRPO requires stochastic exploration through multiple trajectory samples, where policy updates depend on the probability distribution of trajectories and their associated reward signals. However, flow matching models utilize deterministic ODE sampling:
\begin{equation}
    \mathrm{d}\mathbf{x}_t=\mathbf{v}_t\mathrm{d}t,
\end{equation}
where $v_t$ denotes the velocity field. Flow-GRPO and DanceGRPO convert the sampling process of rectified flows from a deterministic ordinary differential equation (ODE) to an equivalent stochastic differential equation (SDE), ensuring that the marginal probability density function at all time steps remains consistent with that of the original model. The SDE sampling process can be formulated as follows:
\begin{equation}
    \mathrm{d}\mathbf{x}_t=\left(\mathbf{v}_{t}-\frac{1}{2}g_t^2\nabla\log p_t(\mathbf{x}_{t})\right)\mathrm{d}t+g_{t}\mathrm{d}\mathbf{w},
\end{equation}
where $\mathrm{d}\mathbf{w}$ denotes Brownian motion, $\nabla\log p_t(\mathbf{x}_t)$ represents the score function at time $t$, and $g(t)$ is the standard deviation of the noise. 
% This SDE ensures that the marginal probability density function at all time steps remains consistent with that of the original model. 
The final update rule is given by:
\begin{equation}
    \mathbf{x}_{t+\Delta t} = \mathbf{x}_t + \left[ \mathbf{v}_{\theta}(\mathbf{x}_t, t) + \frac{\sigma_t^2}{2t} \left( \mathbf{x}_{t} + (1 - t) \mathbf{v}_{\theta}(\mathbf{x}_t, t) \right) \right] \Delta t + \sigma_t \sqrt{\Delta t} \, \mathbf{\epsilon},
\end{equation}
where $\epsilon \sim \mathcal{N}(0, I)$ is used to inject stochasticity, and $\sigma_t = a \sqrt{\frac{t}{1 - t}}$. Notably, previous GRPO methods typically use a KL regularization term to prevent reward over-optimization. 
% However, our experimental results demonstrate that this term has little effect on model performance. By the way, our LayerTuning-RL significantly mitigates this problem as well.
In our experiment, we demonstrate the promising performance of LayerTuning-RL on this problem without KL regularization, which provides an alternative solution to the problem of reward over-optimization.

\subsection{Dynamic-TreeRPO}
Typically, previous RL approaches generate multiple independent trajectories from the same initial noise to facilitate exploration. However, this practice suffers from two major drawbacks: computational redundancy and limited exploration space. Even with SDE applied at each step, the resulting trajectories often exhibit high similarity. To address this, we construct a binary tree-structured search space, where trajectories share a common prefix before branching into distinct subsequent paths. 
% Modeling sequence generation as a binary tree search process is not only feasible but also highly advantageous. 
Modeling sequence generation as a binary tree search process is more than feasible and highly advantageous to efficiency.
% By calculating and storing the shared prefix once and amortizing the computational cost over it, Dynamic-TreeRPO effectively avoids redundant forward computations.
By calculating and storing the shared prefix once, Dynamic-TreeRPO can effectively avoid redundant computations for the calculation of descendant trajectories.
% Within tree structure, each branching point can be considered an ideal pivot for uncertainty-based exploration, enabling efficient and targeted expansion of inference paths. 
Within tree structure, each branching node can be considered a pivot for exploration, enabling efficient and controllable expansion of inference paths. 
Furthermore, we introduce different noise intensities into each tree layer to enhance intra-group diversity and thus expand the searching space.

As shown in Figure \ref{fig:compare}, we follow MixGRPO to combine SDE and ODE sampling. For a tree of depth $d$, we put it in the sliding window across the denoising time range $S_{\mathrm{tree}} = [\tau, \tau+d]$, where $\tau\in[0, T-d]$ and $T$ is the total number of denoising steps. During the denoising process, SDE sampling is employed within the tree, while ODE sampling is used outside the tree. The path from the initial time step to the root node is computed only once, which significantly reducing the forward propagation overhead. To further enhance the diversity of trajectories within each group, we introduce a differentiable noise intensity function at $k$-th tree layer:
\begin{equation}
    g_{t}(k) = g_{t} \times (1 + \beta \frac{k}{d}),   \label{eq6}
\end{equation}
where $\beta$ is a hyperparameter controlling the rate of noise growth, and $\frac{k}{d}$ is the normalized depth, ensuring that the noise intensity increases linearly along the depth of tree. In Dynamic-TreeRPO, the combined ODE and SDE sampling within the tree can be formulated as:
\begin{equation}
    \mathrm{d}\mathbf{x}_t = 
\begin{cases}
\left( \mathbf{v}_{t} - \frac{1}{2}g^{2}_t\nabla\log p_{t}(\mathbf{x}_t) \right) \mathrm{d}t + g_{t}(k) \mathrm{d}\mathbf{w}, & \text{if } t \in S_{tree} \\
\mathbf{v}_{t} \mathrm{d}t, & \text{otherwise}
\end{cases}
\end{equation}
and thus the denoising update can be optimized as:
\begin{equation}
    \mathbf{x}_{t+\Delta t} = 
\begin{cases}
\mathbf{x}_t + \left[\mathbf{v}_{\theta}(\mathbf{x}_{t}, t) + \frac{\sigma_t^2}{2t} \left( \mathbf{x}_t + (1 - t) \mathbf{v}_{\theta}(\mathbf{x}_{t}, t) \right) \right] \Delta t + \sigma_t \sqrt{\Delta t} \, \mathbf{\epsilon}, & \text{if } t \in S_{tree} \\
\mathbf{x}_t + \mathbf{v}_{\theta}(\mathbf{x}_{t}, t) \Delta t, & \text{otherwise}
\end{cases}
\end{equation}
Finally, the training objective is written as:
\begin{equation}
\begin{split}
    \mathcal{J}(\theta) = \mathbb{E}_{c\sim C,{\substack{\{\mathbf{x}_i\}_{i=1}^{2^{d-1}}}   \
 \sim \pi_{\theta_{\text{old}}}(\cdot|c)}} \bigg[ \frac{1}{2^{d-1}} \sum_{i=1}^{2^{d-1}} \frac{1}{|S_{tree}|} \sum_{t=\tau}^{\tau+d} \\ \min \bigg( \mathbf{\rho}_{t,i} A_i, \text{clip}\big( \mathbf{\rho}_{t,i}, 1-\varepsilon, 1+\varepsilon \big) A_i \bigg) \bigg],
\end{split}
\end{equation}

where the policy ratio $\mathbf{\rho}_{t,i}$ and advantage function $A_i$ are consistent to previous settings~\citep{li2025mixgrpo}. 
Intuitively, the concept of sampling group is determined by the trajectories of binary tree. Compared with previous methods, our optimization is performed only in the tree, rather than across all time steps. 
% The number of function evaluations (NFE) for ${\pi_{\theta_{old}}}$ is reduced from $2^{d-1} \times T$ to $T_n + 2^{d-1} \times (T - T_n)$, where $T_n$ denotes the time step of the tree root node. 
The number of function evaluations (NFE) for ${\pi_{\theta_{old}}}$ is reduced from $2^{d-1} \times T$ to $\tau + 2^{d-1} \times (T - \tau)$. 
The pseudo-code of our method can be found in Algorithm~\ref{dynamic_algorithm}. 
In summary, we achieve significant improvements on both computational efficiency and exploration diversity through the dynamic weighted tree-structured sampling approach.

\begin{algorithm}
\caption{DYNAMIC-TREERPO Training Algorithm}
\label{dynamic_algorithm}
\renewcommand{\algorithmicrequire}{\textbf{Input:}}
\renewcommand{\algorithmicensure}{\textbf{Output:}}
\begin{algorithmic}[1]
\REQUIRE Policy model $\pi_\theta$, reward models $\{R_k\}_{k=1}^K$, prompt dataset $\mathcal{C}$, total sampling steps $T$, Tree depth $d$. 
\ENSURE
    Optimized policy model $\pi_\theta$
\FOR{training iteration $m = 1$ to $M$}
    \STATE Sample batch prompts $\mathcal{C}_b \sim \mathcal{C}$
    \STATE Update old policy: $\pi_{\theta_{\text{old}}} \gets \pi_\theta$ 
    % \COMMENT{$\pi_{\theta_{\text{old}}}$ mixed sampling loop}
    \FOR{each prompt $\mathbf{c} \in \mathcal{C}_b$}
        \STATE Generate $2^{d-1}$ samples: $\{\mathbf{o}_i\}_{i=1}^{2^{d-1}} \sim \pi_{\theta_{\text{old}}}(\cdot|\mathbf{c})$ with tree structured sampling
        \STATE Compute rewards $\{r_i^k\}_{i=1}^G$ using $R_k$
        \STATE Find index of sample with maximum advantage: $i^* = \arg\max \{r_i^k\}_{i=1}^G$ 
        \FOR{each sample $i = 1$ to $2^{d-1}$}
            \FOR{sampling timestep $t=0$ \textbf{to} $T-1$} 
                \IF{$t \in \texttt{Tree}$}
                    \STATE Use ree structured Sampling with  to get $\mathbf{x}^{i}_{t+1}$  with different noise intensities
                \ELSE
                    \STATE Use ODE Sampling to get $\mathbf{x}^{i}_{t+1}$
                \ENDIF
                \STATE Calculate multi-reward advantage: $A_i \gets \sum_{k=1}^K \frac{r_i^k - \mu^k}{\sigma^k}$ 
            \ENDFOR
        \ENDFOR
        
        \FOR{each timestep {$t \in \texttt{Tree}$}}
            \STATE Update policy via gradient ascent: $\theta \gets \theta + \eta \nabla_\theta \mathcal{J}(r_{i_*}^k)$
        \ENDFOR
    \ENDFOR
     \STATE Update {Tree} root node position
\ENDFOR
\end{algorithmic}
\end{algorithm}

\subsection{Dynamic-Adaptive Clipping Bounds for Trajectory Reward}
Clipped probability ratio has been widely adopted in reinforcement learning to stabilize policy updates and prevent excessively large and unstable parameter changes during optimization. By introducing a hyperparameter $\epsilon$, the probability ratio is constrained within fixed clipping boundaries, as shown in the following equation:
\begin{equation}
    |\mathbf{\rho}_{t,i}  -1|= |\frac{\pi_{\theta}(\mathbf{x}_{t-1,i}|\mathbf{x}_{t,i},c)}{\pi_{\theta_{old}}(\mathbf{x}_{t-1,i}|\mathbf{x}_{t,i},c)} -1| \leq  \varepsilon.
\end{equation}
For flow matching models, this constraint implies that an identical restriction is imposed on the relative change of policy outputs at every time steps. 
Such a "one-size-fits-all" strategy suffers from a fundamental drawback that it arbitrarily restricts the exploration with low-probability without considering its potentiality.
% potentially crucial time steps, neglecting their absolute probability basis and thus suppressing the model's ability to learn new knowledge and discover diverse reasoning paths. 
Therefore, the capability of learning new knowledge and discovering diverse reasoning paths may be suppressed under this setting. 
% Although some methods have been proposed in the text generation domain to address this issue, such as DAPO, which uses asymmetric clipping boundaries, and DCPO, which dynamically associates the token's own probability with the clipping boundary, these approaches either retain fundamental limitations or are not applicable to the T2I domain.
Some methods have been proposed in the text generation domain to address this kind of issue.
For example, DAPO uses asymmetric clipping boundaries, and DCPO dynamically associates the token's own probability with the clipping boundary.
However, these approaches still retain some fundamental limitations or even inapplicable to the T2I task.
To this end, we introduce the reward value of each trajectory into the clipping boundary, as shown in the following equation:
\begin{equation}
    \varepsilon_t = \varepsilon_{low} + (\varepsilon_{high} -  \varepsilon_{low})\cdot e^{-\eta R_{(i)}}, \label{eq11}
\end{equation}
where $R_{(i)}$ is the reward value corresponding to the trajectory of the $i$-th leaf node where $i \in [0, 2^{d-1}]$, $\eta$ is the reward sensitivity factor, and $\varepsilon_{\mathrm{low}}, \varepsilon_{\mathrm{high}}$ are the lower and upper clipping thresholds pre-set following DAPO. Equation \ref{eq11} allows the clipped probability ratio to be dynamically adjusted within $[\varepsilon_{\mathrm{low}}, \varepsilon_{\mathrm{high}}]$ according to the trajectory's reward value, permitting bolder exploration steps in low-reward regions while adopting more cautious fine-tuning in high-reward regions.

\subsection{LayerTuning-RL}
Current training paradigms of T2I models employ RL or SFT approaches. 
Generally, SFT method enables the model to learn expert-level reasoning trajectories, while RL allows the model to autonomously explore and select the aligned inference paths within the existing knowledge.
% However, sequentially applying SFT followed by RL often results in catastrophic forgetting. 
Some works sequentially employ SFT and RL methods, which often result in catastrophic forgetting.
To better integrate the advantages of both paradigms, we propose LayerTuning-RL, a tightly coupled framework combining supervised-tuning and MIXTree-GRPO. 
Specifically, we introduce the following supervised-tuning objective at each time step $t$:
\begin{equation}
    \mathcal{J}(\theta)_{\mathrm{SFT}}  = \mathbb{E} \left[ \| V^*_t- V^\theta_{t,i} \|_2^2 \right], \label{eq12}
\end{equation}
where $V^*_t = V^\theta_{t, \arg\max(R)}$ denotes the best predicted trajectory at time step $t$, which obtains the highest reward score in the set of $V^\theta_{t,i}$.
% representing the optimal optimization direction at step $i$. 
In the optimization of tree nodes, $V^*_i$ is used as the pseudo target of supervised-tuning objective, which provides the guidance for the optimization directions of each layer in one sampling group.
% PRM guidance for the optimization directions of other samples in the same group.
% The role of supervised-tuning in this process is to help the model directly align with the optimal result at any step.
To some extent, this supervised-tuning objective can be considered as PRM for each layer in RL paradigms, and thus construct the paradigm of our LayerTuning-RL.
% Meanwhile, RL further optimizes the model based on the advantage, serving as an auxiliary optimization mechanism. 
% During training, supervised-tuning and RL interact, enabling better collaboration. We propose the following fusion objective:
The overall training objective of LayerTuning-RL can be written as:
% \begin{equation}
% \begin{split}
%      \mathcal{J}(\theta)_{\text{fusion}} = \mathbb{E}_{c\sim C,{\substack{\{\mathbf{x}_i\}_{i=1}^{2^{d-1}}}   \
%  \sim \pi_{\theta_{\text{old}}}(\cdot|c)}} \bigg[ \frac{1}{2^{d-1}} \sum_{i=1}^{2^{d-1}} \frac{1}{|S_{tree}|} \sum_{t=\tau}^{\tau+d} \min  \\ \bigg( \mathbf{\rho}_{t,i} A_i, \text{clip}\big( \mathbf{\rho}_{t,i}, 1-\varepsilon_t, 1+\varepsilon_t \big) A_i \bigg)+\lambda \times \| V^*_i- V^\theta_{t,i} \|_2^2 \bigg]
% \end{split}
% \end{equation}
\begin{equation}
\begin{split}
     \mathcal{J}(\theta)_{\text{fusion}} = \mathcal{J}(\theta) + \lambda \times \mathcal{J}(\theta)_{\mathrm{SFT}},  \label{eq13}
\end{split}
\end{equation}
where $\lambda$ denotes the hyperparameter for fusing the SFT and RL paradigms.

LayerTuning-RL is implemented as a collaborative framework, where RL provides overall trajectory-level supervision based on reward optimization, and supervised-tuning offers finer-grained supervision at each node.
We claim that the design of LayerTuning-RL offers three main advantages: (1) it avoids catastrophic forgetting caused by two-stage training; (2) it improves path exploration efficiency through PRM-based supervised-tuning guidance; and (3) it guarantees the performance improvement of RL by strategically transferring beneficial knowledge from supervised-tuning to LayerTuning-RL.
More details can be discussed in experiments.

\section{EXPERIMENTS}
\subsection{EXPERIMENT SETUP}
\noindent\textbf{Dataset.} We evaluate Dynamic-TreeRPO on the HPDv21 dataset~\citep{wu2023human}. The training set contains 103,700 prompts.
However, we achieve state-of-the-art performance using only a small fraction of the training data in practice. The test set consists of 400 prompts.

\noindent\textbf{Evaluation Metrics.}  We align with MixGRPO and assess performance on two metrics: computational cost and generation quality.
To measure the computational cost, we report the number of function evaluations (NFE) and the average training time per GRPO iteration, which faithfully reflect the actual training overhead. For quality assessment, we employ three reward models, including HPS-v2.1, PickScore~\citep{kirstain2023pick}, ImageReward~\citep{xu2023imagereward}, as our evaluators.
% Our evaluation protocol aligns with that of MixGRPO, assessing performance along two key dimensions: computational overhead and generation quality. Regarding computational overhead, we adopt the number of function evaluations (NFE), which is further decomposed into o $\text{NFE}_{\pi_{\theta_{\text{old}}}}$ and $\text{NFE}_{\pi_{\theta}}$. $\text{NFE}_{\pi_{\theta_{\text{old}}}}$  represents the number of forward propagation of the reference model for computing the policy ratio and generating images. $\text{NFE}_{\pi_{\theta}}$ is the number of forward propagation of the policy model solely for the policy ratio. In addition, we report the average training time per GRPO iteration to provide a realistic and practical measure of training efficiency. For performance evaluation, four reward models—HPS-v2.1, PickScore, ImageReward, and Unified Reward—are adopted as the final assessment metrics.

\noindent\textbf{Implementation Details.}  We use FLUX.1-dev as the base model, an advanced text-to-image diffusion model based on flow matching. The trajectory tree is configured with depth $d = 4$, yielding $8$ leaf nodes. The noise growth magnitude $\beta$ is set to $0.7$, and the reward sensitivity factor $\eta$ for clipping probability ratios is set to $0.5$. The clipping thresholds $\varepsilon_{\text{low}}$ and $\varepsilon_{\text{high}}$ are set to $5 \times 10^{-5}$ and $5 \times 10^{-3}$, respectively. The fusion coefficient $\lambda$, which balances supervised fine-tuning and the reinforcement learning paradigm, is set to $0.02$.

During training, for each prompt, we generate $8$ images (one per leaf node), and each image is sampled by $T = 25$ denoising steps. The model is fine-tuned for $100$ iterations on $8$ NVIDIA H100 GPUs with a global batch size of $16$. We use the AdamW optimizer~\citep{loshchilov2017decoupled} with a learning rate of $5 \times 10^{-6}$ and weight decay $1 \times 10^{-4}$. Training is conducted in \texttt{bfloat16} mixed precision. All other hyper-parameters are kept consistent to those of MixGRPO and other baselines.

\begin{figure*}[h]
    \centering
    \includegraphics[width=0.99\textwidth]{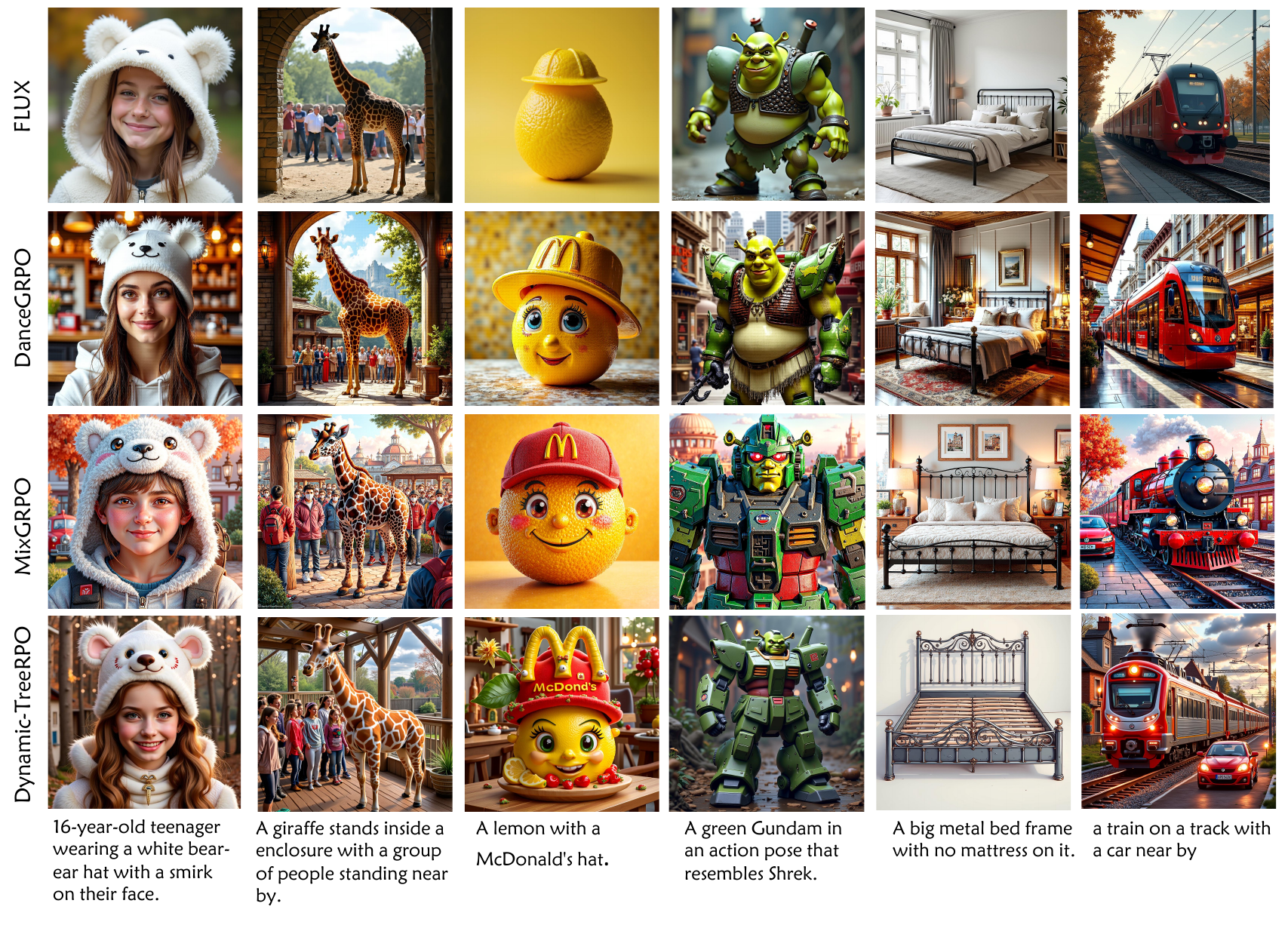}
    \caption{Qualitative comparison. Dynamic-TreeRPO achieves superior performance compared to Flux,DanceGRPO and MixGRPO in terms of semantics, aesthetics and text-image alignment.}
    \label{fig:qualitative}
\end{figure*}

% \begin{table}[h]
%   \centering
%   \caption{Comparison results of computational efficiency and image quality. Dynamic-TreeRPO achieves the best performance across multiple metrics, with the top result in each column highlighted in \textbf{bold}. ${\text{NFE}}_{\pi_{\theta_{{\text{old}}}}}$represents the number of forward propagation of the reference model for computing the policy ratio and generating images. ${\text{NFE}}_{\pi_{\theta}}$ is the number of forward propagation of the policy model solely for the policy ratio. In Dynamic-TreeRPO, we report the average NFE per sample.}
%   \resizebox{\linewidth}{!}{
%     \input{Tables/main_multi_reward.tex}
%   }
%   \label{tab:main_multi_reward}
%   \vspace{-1.0em}
% \end{table}

\subsection{MAIN EXPERIMENTS} 
% We conducted a comprehensive evaluation of computational efficiency and generation quality in Table ~\ref{tab:main_multi_reward}. Dynamic-TreeRPO achieves superior quality while significantly reducing training costs, attaining the highest scores across all three metrics: HPS-v2.1, PickScore, and ImageReward. Specifically, HPS-v2.1 improves from 0.367 (MixGRPO) to 0.385, PickScore increases from 0.237 to 0.251, and ImageReward rises from 1.629 to 1.770. Meanwhile, the average training iteration time is reduced to 151 seconds, which is 9\% faster than MixGRPO-Flash. We also performed qualitative analysis: Figure~\ref{fig:qualitative} visually compares our method with FLUX, DanceGRPO, and MixGRPO, demonstrating that Dynamic-TreeRPO generates richer details and exhibits stronger semantic alignment, with noticeable improvements in instruction following and aesthetic quality. Furthermore, the reward curves in Figure~\ref{fig:compare} provide additional evidence of our method’s superiority, showing faster stabilization with fewer timesteps and achieving higher reward levels at earlier stages.

We conduct comprehensive experiments to evaluate the training efficiency and generation quality of our Dynamic-TreeRPO. In the left part of Figure \ref{fig:compare},  the reward curve of our method consistently surpasses those of MixGRPO and DanceGRPO, indicating a faster stabilization and higher reward levels in earlier stages. The training effectiveness of our method is also evidenced in Table \ref{tab:main_multi_reward}, where it achieves an average training iteration time of 151 seconds, a reduction of $114\%$ compared to DanceGRPO. More importantly, our method also achieves impressive qualitative generation results. This is evident in Figure \ref{fig:qualitative}, which shows that our approach exhibits a stronger ability for semantic alignment (\textit{e.g.}, ``Gundam resembles Shrek'' of column 4, and not vice versa) and instruction following (e.g., ``no mattress on it" of column 5). As presented in Table \ref{tab:main_multi_reward}, our method significantly surpasses the four baseline methods on all three evaluation metrics.

\begin{table}[h]
  \centering
  \caption{Comparison results of computational efficiency and image quality. Dynamic-TreeRPO achieves the best performance across multiple metrics, with the top result in each column highlighted in \textbf{bold}. In Dynamic-TreeRPO, we report the average NFE per sample.}
  \resizebox{\linewidth}{!}{
    \begin{tabular}{lccccccc}
\toprule[1pt]
\multirow{2}{*}{\textbf{Method}} & 
\multirow{2}{*}{\textbf{$\text{NFE}_{\pi_{\theta_{\text{old}}}}$}} & \multirow{2}{*}{\textbf{$\text{NFE}_{\pi_{\theta}}$}} &
\multirow{2}{*}{\textbf{Iteration Time~(s)}$\downarrow$}  & 
\multicolumn{4}{c}{\textbf{Human Preference Alignment}} \\
\cmidrule(lr){5-8}
& & & & \textbf{HPS-v2.1}$\uparrow$ & 
\textbf{Pick Score}$\uparrow$ & 
\textbf{ImageReward}$\uparrow$ & \\
\midrule

FLUX& / & / & / & 0.313 & 0.227 & 1.088  \\
% \midrule
% OnlineDPO& - & 25 & 0 & - & - & - & - \\
\midrule
\multirow{2}{*}{DanceGRPO}
& 25 & 14  & 323   & 0.356 & 0.233 & 1.436  \\
& 25 & 4   & 241   & 0.334 & 0.225 & 1.335  \\
\midrule
MixGRPO & 25 & 4 & 240 & {0.367} & {0.237} & {1.629}  \\
MixGRPO-Flash & 16 & 4 & 166 & {0.358} & {0.236} & {1.528}  \\
\midrule
Dynamic-TreeRPO & 13.8(Avg) & 4 & \textbf{151} & \textbf{0.385} & \textbf{0.251} & \textbf{1.770}  \\
\bottomrule[1pt]

% \multirow{2}{*}{\name{}-Flash}
% & 16 (Avg) & 4 & \underline{112.372}  & \underline{0.358} & \underline{0.236} & 1.528  \\
% & 8 & 4$^*$ & \textbf{83.278} & 0.357 & 0.232 & \underline{1.624}  \\
% \bottomrule
\end{tabular}

  }
  \label{tab:main_multi_reward}
  \vspace{-1.0em}
\end{table}

\subsection{ABLATION EXPERIMENTS} 
We have meticulously designed a series of experiments. Table \ref{tab:abl_mian} presents a comprehensive ablation study of the proposed Dynamic-TreeRPO, Dynamic Clipping, and LayerTuning-RL components. The results demonstrate that each component is individually effective. Specifically, introducing Dynamic-TreeRPO on top of the baseline effectively reduces training cost while maintaining performance across multiple evaluation metrics. Incorporating LayerTuning-RL and Dynamic Clipping addresses issues of inefficient exploration and training instability, enabling faster reward growth and smoother convergence.
% \begin{table}[h]
%   \centering
%   \caption{Ablation experiments of Dynamic-TreeRPO. Blue shows performance gain over the baseline (the first row). }
%   \resizebox{\linewidth}{!}{
%     \input{Tables/ablation_main.tex}
%   }
%   \label{tab:ablation_main}
%   \vspace{-1.0em}
% \end{table}

%To further investigate the impact of the noise growth hyperparameter $\beta$ on the exploration trajectory, We conduct experiments based on baseline with Dynamic-TreeRPO, Table~\ref{tab:abl_beta} presents the performance under different $\beta$ values. We observe that setting $\beta$=0.3 overly constrains exploration, leading to suboptimal final rewards. Conversely, $\beta$=0.9 results in excessive noise intensity variation, causing training instability. A value of $\beta$=0.7 strikes the optimal balance between exploration and stability, achieving the highest reward while ensuring consistent convergence. 

To investigate the effect of the noise growth parameter $\beta$ in Equation \ref{eq6} on the performance of our Dynamic-TreeRPO, we sampled a range of values evenly spaced between $0.3$ and $0.9$. The final results corresponding to these values of $\beta$ are summarized in Table \ref{tab:abl_beta}. We observed that all three metrics initially exhibit a positive correlation with $\beta$, peaking at $\beta=0.7$ before declining.

\begin{table}[ht]
\centering
\scriptsize
\setlength{\tabcolsep}{1pt}

\begin{minipage}[t]{0.46\textwidth}
    \centering
    \caption{Ablation experiments of Dynamic-TreeRPO}
    \label{tab:abl_mian}
    \vspace{0.3em}
    \renewcommand\arraystretch{1.2}
\tabcolsep=0.1cm
\begin{tabular}{l|c|c|c}

\toprule[1pt]

$ $ Component & $ $ HPS-v2.1$\uparrow$ $ $ & $ $ Pick Score$\uparrow$ $ $ & $ $ ImageReward$\uparrow$ $ $  \\
\midrule [1pt]
$ $ baseline &  0.313 & 0.227 & 1.088  \\

% \midrule
$ $ +Dynamic-Tree&  0.361 & 0.238 & 1.591  \\

% \midrule
$ $ +Dynamic-Clipping $ $ &  0.369 & 0.242 & 1.674  \\
% \midrule
$ $ +LayerTuning-RL & 0.385 & 0.251 & 1.770  \\
\bottomrule [1pt]
\end{tabular}

\end{minipage}%
\hfill
\begin{minipage}[t]{0.46\textwidth}
    \centering
    \caption{Comparison for noise growth hyperparameter $\beta$}
    \label{tab:abl_beta}
    \vspace{0.3em}
    \renewcommand\arraystretch{1.2}
\tabcolsep=0.1cm
\begin{tabular}{l|c|c|c}
\toprule[1pt]
$  $  $\beta$ &$  $ HPS-v2.1$\uparrow$ $  $ &$  $ Pick Score$\uparrow$ 
 $  $ &$  $ ImageReward$\uparrow$ $  $  \\
\midrule[1pt]
$  $ 0.3 $  $ &  0.348 & 0.221 & 1.337  \\

% \midrule
$  $ 0.5 $  $ & 0.351 & 0.228 & 1.448  \\

% \midrule
$  $ 0.7 $  $& 0.361 & 0.238 & 1.591  \\

% \midrule
$  $ 0.9 $  $& 0.355 & 0.229 & 1.611  \\

\bottomrule[1pt]
\end{tabular}
    
\end{minipage}

\vspace{-1.0em}
\end{table}

\begin{figure*}[h]
    \centering
    \includegraphics[width=0.99\textwidth]{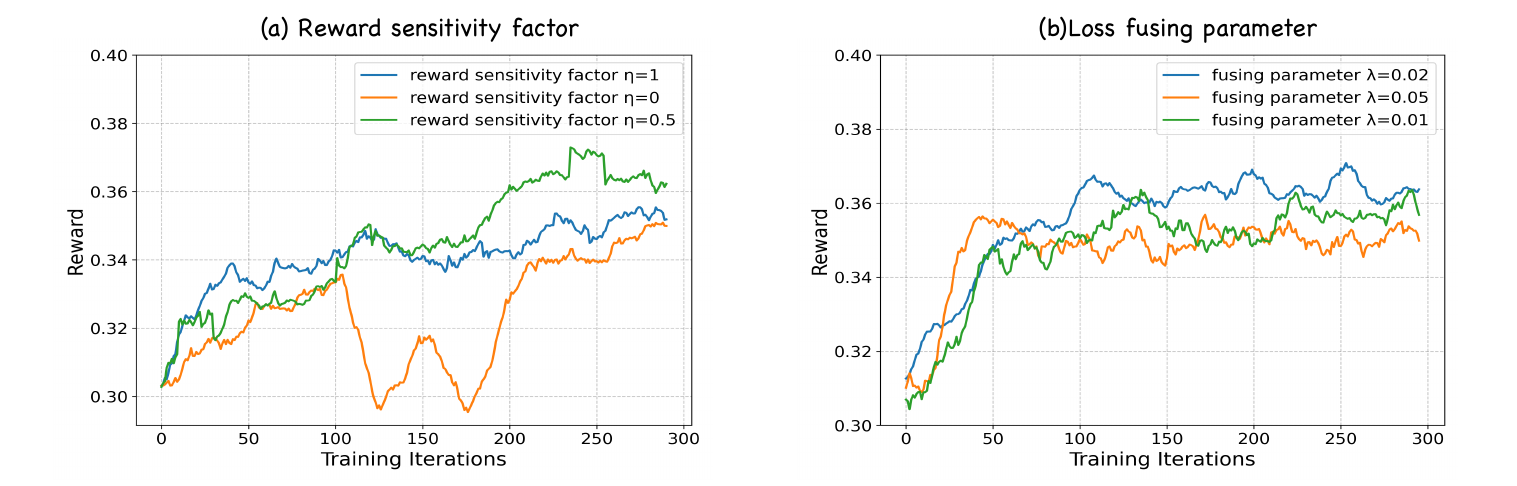}
    \caption{Ablation Studies on  reward sensitivity factor and balancing parameter in LayerTuning-RL.}
    \label{fig:alation_2}
\end{figure*}

Furthermore, we conduct experiments to investigate the impact of the reward sensitivity factor $\eta$ in Equation \ref{eq11} and the balancing parameter $\lambda$ in Equation \ref{eq13}. As shown in Figure~\ref{fig:alation_2}(a), under the baseline + Dynamic Clipping framework, the setting $\eta$=0, which corresponds to the maximum clipping of the probability ratio, leads to noticeable instability during training, as reflected by the oscillatory learning curve. In contrast, the setting $\eta$=0.5 yields stable training dynamics while preserving sufficient exploration, indicating that the majority of generated samples effectively contribute to model updates. Figure~\ref{fig:alation_2}(b) validates the effect of the balancing parameter $\lambda$.  A larger $\lambda$ allows faster convergence in the early training stage, but imposes limitations in the later phase. Therefore, we set $\lambda$=0.02, which improves convergence without incurring additional computational cost and ensures the performance of RL through PRM.

% Figure~\ref{fig:alation_2}(a) illustrates the impact of the reward sensitivity factor $\eta$ under varying pruning probability ratios. We set  $\eta=0.5$,  which maintains training stability. Figure~\ref{fig:alation_2}(b) validates the balancing parameter $\lambda$ in LayerTuning-RL. A larger $\lambda$ enables faster convergence in the early training stages but imposes limitations in later phases. We therefore set $\lambda$=0.02 , which improves convergence without incurring additional computational cost and ensures the performance of RL through supervision from the PRM.

\section{CONCLUSION}
In this paper, we introduce Dynamic-TreeRPO, a RL framework built upon a sliding tree structure. By incorporating dynamically adaptive clipping boundaries constrained by reward signals and integrating a training strategy combined with SFT, our approach accelerates training convergence while enabling efficient exploration of the search space. It effectively mitigates common issues in GRPO, such as catastrophic forgetting and inefficient exploration. 
% Extensive experimental results demonstrate that our method achieves superior semantic consistency and visual quality in image generation.

\clearpage
\bibliography{iclr2026_conference}
\bibliographystyle{iclr2026_conference}
\clearpage
\end{document}